\documentclass[conference]{IEEEtran}

\usepackage{flushend}
\usepackage{epsfig}
\usepackage{verbatim}
\usepackage{url}
\usepackage{cite}

\begin{document}

\title{Depth-Map Generation using Pixel Matching in Stereoscopic Pair of Images}

\author{\IEEEauthorblockN{Asra Aslam}
\IEEEauthorblockA{Insight Centre for Data Analytics\\ National University of Ireland \\ Galway, Ireland\\
\textit{Email: asra.aslam.7@gmail.com}}
\and
\IEEEauthorblockN{Mohd. Samar Ansari}
	\IEEEauthorblockA{Software Research Institute \\ Athlone Institute of Technology \\ Athlone, Ireland\\
		\textit{Email: mdsamar@gmail.com}}
}
\maketitle

\begin{abstract}
Modern day multimedia content generation and dissemination is moving towards the presentation of more and more `realistic' scenarios. The switch from 2-dimensional (2D) to 3-dimensional (3D) has been a major driving force in that direction.
Over the recent past, a large number of approaches have been proposed for creating 3D images/videos most of which are based on the generation of depth-maps. This paper presents a new algorithm for obtaining depth information pertaining to a depicted scene from a set of available pair of stereoscopic images. The proposed algorithm performs a pixel-to-pixel matching of the two images in the stereo pair for estimation of depth. It is shown that the obtained depth-maps show improvements over the reported counterparts.
\end{abstract}


\IEEEpeerreviewmaketitle

\section{Introduction}

Over the recent past, there has been a tremendous increase in the amount of multimedia data generated and disseminated across the globe \cite{shifa2019lightweight,aslam2018towards}. This has necessitated active research in the fields of image and video processing, with applications and solution being continuously developed for areas like surveillance and security, crowd management, healthcare \cite{aslam2015improved}, computer vision, and many others. 

The advent of digital cameras has revolutionized the way users take pictures. Compared to their analog counterparts, such digital cameras (including the ones present in mobile devices) provide fast and easy image procurement, storage and retrieval. However, most of the popular digital cameras are capable of capturing a 2-dimensional (2D) projection of the scene while the components corresponding to depth are lost (not recorded). 3-dimensional (3D) imaging has emerged as an advancement to the conventional 2D technology with the additional information of depth included. 3D cameras have started to appear in the market but are prohibitively expensive. For a normal user with a 2D digital camera, 3D images may be constructed by extracting the depth information from 2D images using a variety of techniques proposed over the past \cite{battiato2004depth,sun2003stereo,fahmy2013stereo,kamencay2011stereo,birchfield1998pixel}.
Among these methods, depth-map generation from a stereo pair of images is the most popular one \cite{kanade1994stereo,kamencaynew,yang2013depth,battiato20043d,shultz3d,henry2010rgb,ideses2008depth}. It finds many of its applications in 3D imaging \cite{tam2013generating} , decoding light field images \cite{venkataraman2015systems}, hand tracking \cite{qian2014realtime} etc.

Essentially, a depth-map is a Grey-coded 2D image that gives the perception of depth by the intensity of colors. Darker regions in the Depth-Map are created for signifying that an object is far away and this darker color gradually decreases to brighter with decrease in depth and finally becomes white for closer objects. This paper presents an algorithm for the creation of depth-map starting from a stereo pair of images \textit{i.e} left (L--) and right (R--) images corresponding to the same scene by performing a pixel-to-pixel matching. The algorithm finds matching pixels by comparing the RGB components of the pixels in the L-- and R--images. If the dissimilarity between the compared pixels is found to be less than a pre-specified tolerance (user defined) then those pixels are considered by the algorithm as a `matching pair' of pixels. Binocular disparity is then calculated for the matched pixels which is further utilized to estimate depth information.

The remainder of the paper is arranged as follows. A brief overview of methods for creation of depth-map is presented in Section--\ref{sec:related}. Pertinent theoretical information related to depth-maps and the proposed algorithm are discussed in Section--\ref{sec:depthmap}. Few examples and their derived depth-maps are shown in--\ref{sec:results} along with comparisons with existing works. Lastly, concluding remarks appear in Section-- \ref{sec:conclusion}.

%%%%%%%%%%%%%%%%%%%%%%%%%%%%%%%%%%%%%%%%%%%%%%%%%%       
 
\section{Related Work}
\label{sec:related}

This section discusses various approaches for generation of depth-maps. Some of them are supervised and others are unsupervised. A MATLAB algorithm was developed to construct depth mask using two static images \cite{shultz3d}. The algorithm displays the two images and the user matches corresponding points in both images. From the displacement of the selected image points the algorithm estimates a depth surface for the scene. It is a supervised approach, here user interaction is required for point matching \cite{shultz3d}. An approach which is based on both monocular and stereo cues was proposed for estimating depth \cite{saxena2007depth}. In their work they apply a Markov Random Field (MRF) learning algorithm to capture monocular cues and then combine them with stereo cues to obtain depth maps. However some of these monocular cues are based on prior knowledge, which requires supervised learning \cite{saxena2007depth}. To detect depth discontinuities from a stereo pair of images, an algorithm was presented that matches individual pixels in corresponding scan-line pairs, while allowing occluded pixels to remain unmatched, then propagates the information between scan-lines \cite{birchfield1999depth}. In another approach, high accuracy depth maps  are generated by using structured light. This approach relies on using a pair of cameras and one or more light projectors that cast structured light patterns onto the scene. They have developed a methodology to acquire truth disparity measurements accurately aligned with stereo image pairs \cite{scharstein2003high}. It is difficult to match left and right images by pixel to pixel, so many approaches are proposed which matches the two images by selecting particular size window. A method is presented to select an appropriate window by evaluating the local variation of the intensity and the disparity. This method searches for a window that produces the estimate of disparity with the least uncertainty for each pixel of an image, then embedded this adaptive-window method in an stereo matching algorithm \cite{kanade1994stereo}. In almost all the above mentioned techniques, an exact pixel to pixel matching is \textit{not} performed thereby affecting the quality of the generated depth maps. In this paper a novel algorithm of pixel to pixel matching is proposed, which takes different thresholds into account at different levels. Few results of proposed algorithm are appear in \cite{aslam2014towards}, and it is an improved version over that in terms of removing its shortcomings.

% % % % % % % % % % % % % % % % % % % % % % % % %
\section{Proposed Algorithm}
\label{sec:depthmap}

A depth map is a 2D image that gives the depth (with respect to the viewpoint) of an object as a function of the image coordinates. Usually, it is represented as a Grey level image with the intensity of each pixel registering its depth. The tasks required for creation of Depth-map are: (\textit{i}) Capturing Images, (\textit{ii}) Image Preprocessing, (\textit{iii}) Depth Estimation, and (\textit{iv}) Calculation of color value for all pixels.

\subsection{Capturing Images}
The proposed algorithm uses two images of the same scene. The left and right images can be taken with two digital cameras. They capture images of the same scene, at the same time. These cameras are slightly displaced by some horizontal distance. This horizontal distance should be approximately equal to the spacing between the two eyes. For each set of images the camera position should be carefully controlled with respect to the altitude and viewing vector of the camera.

\subsection{Image Preprocessing}
Read the headers of L-- and R--images to find number of pixels along the height and width of the images. Read both images byte by byte and store them. Separate RGB (Red, Green \& Blue) components of each pixel.

\subsection{Depth Estimation}
Depth estimation is the calculation of depth of different objects in a scene from a multiple views or images. It is required to find corresponding pixels in the different views, i.e., point of correspondence that identifies the same 3D points. By finding these points, the depth information is calculated by following  three main steps: (\textit{1}) Matching of Pixel. (\textit{2}) Choose the best in case of conflicts. (\textit{3}) Disparity Calculation.

\subsubsection{Matching of Pixel}
The matching criterion is based on  Sum of Absolute Differences (SAD). SAD is a matching cost function, the metric of which is calculated for the three color channels and the resulting three absolute difference values are simply added. If this value is less than or equal to $\pm$2.5\% of tolerance, then only consider the corresponding pixel of right image to be the matching pixel of the left image.

\begin{figure}[!htb]
 \begin{center}
 \includegraphics[width=0.5\textwidth]{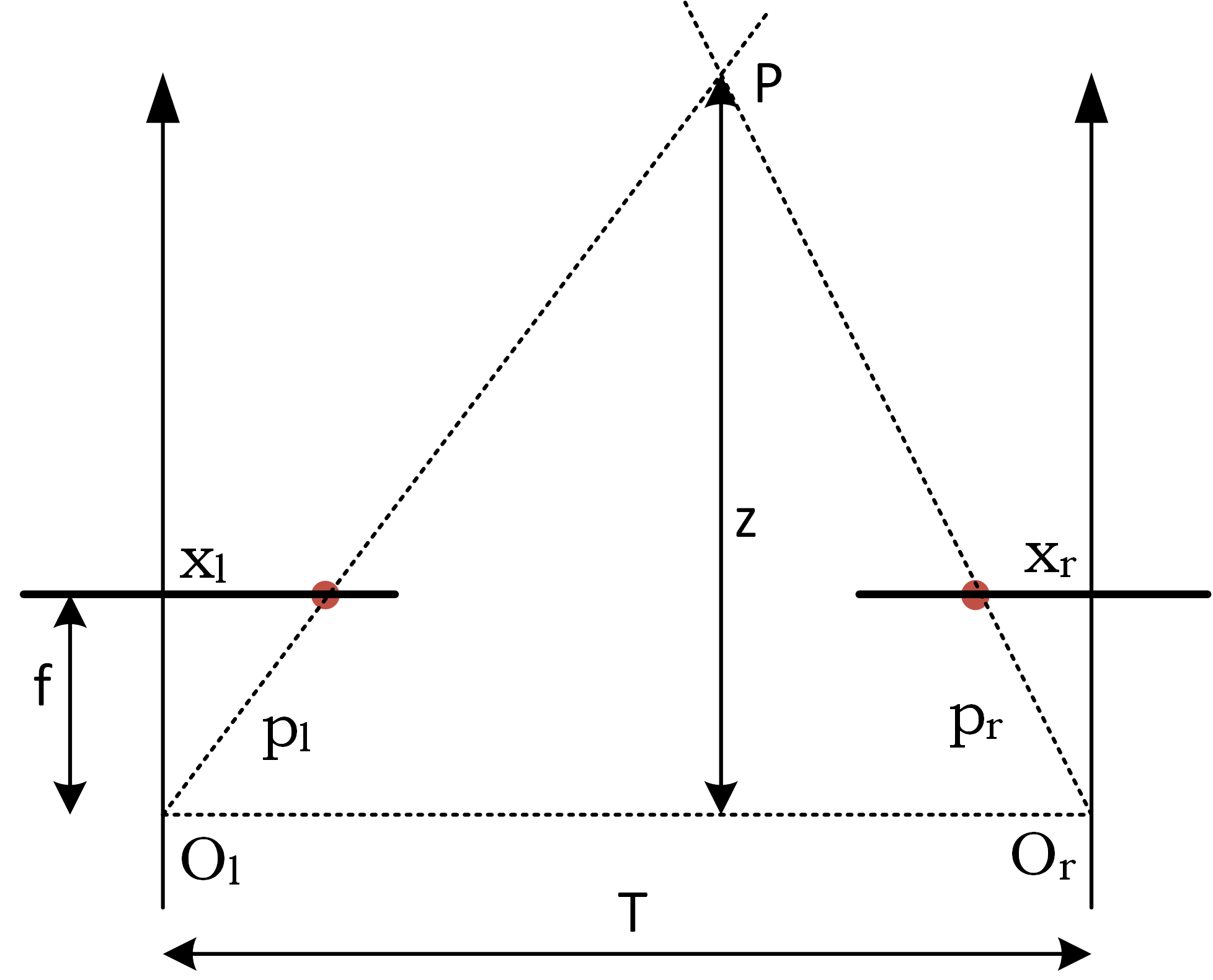}
 \caption{Calculation of Binocular Disparity}
  \label{fig:disparity} 
 \end{center}
\end{figure}
\begin{figure}[htb]
\begin{center}
\includegraphics[width=0.5\textwidth]{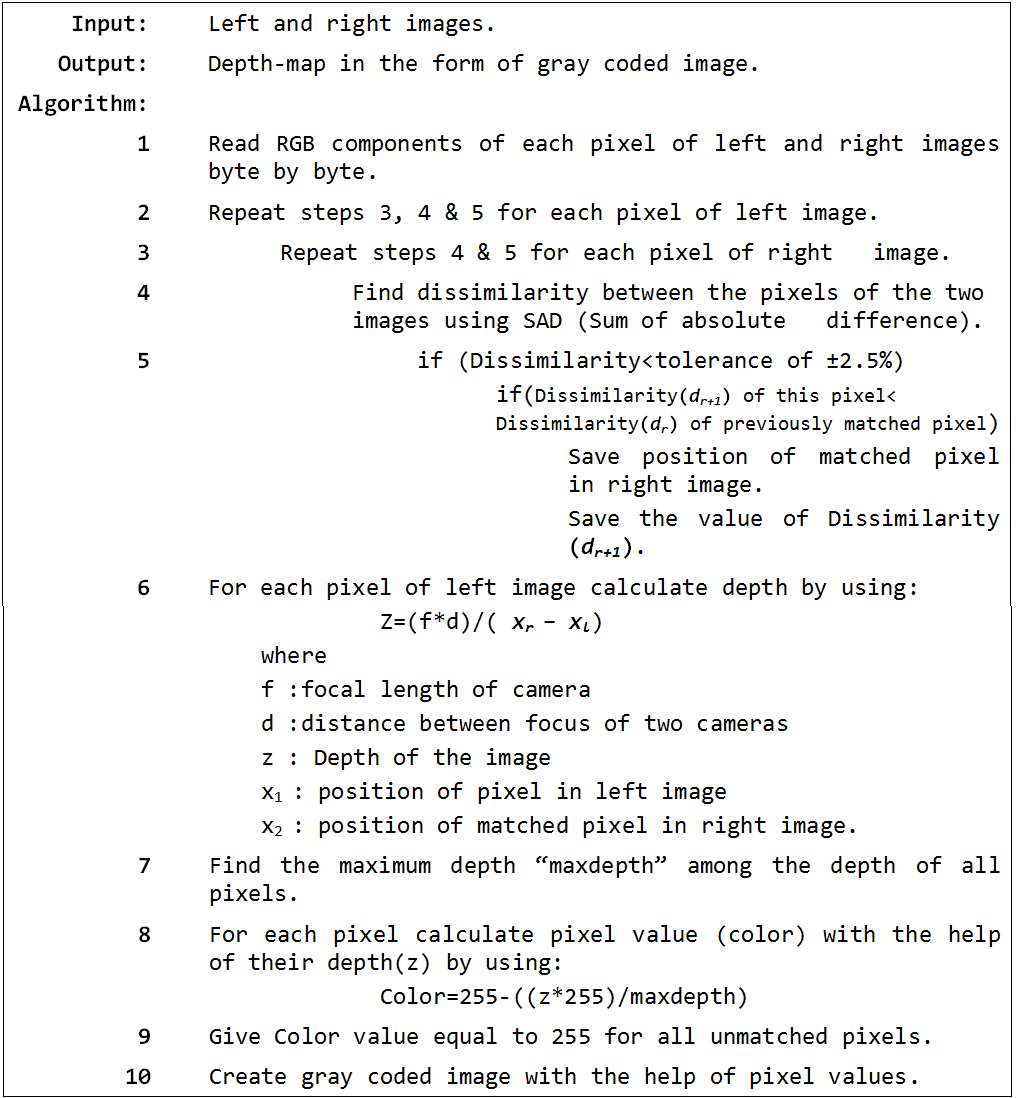}
\end{center}
\caption{Proposed algorithm for depth-map creation}
\label{fig:algo-depthmap}
\end{figure} 

% % % % % % % % % % % % % % % % % % % % % % % % % %

\begin{figure*}[!htb]
	\begin{center}
		\includegraphics[width=0.98\textwidth, height=0.95\textheight]{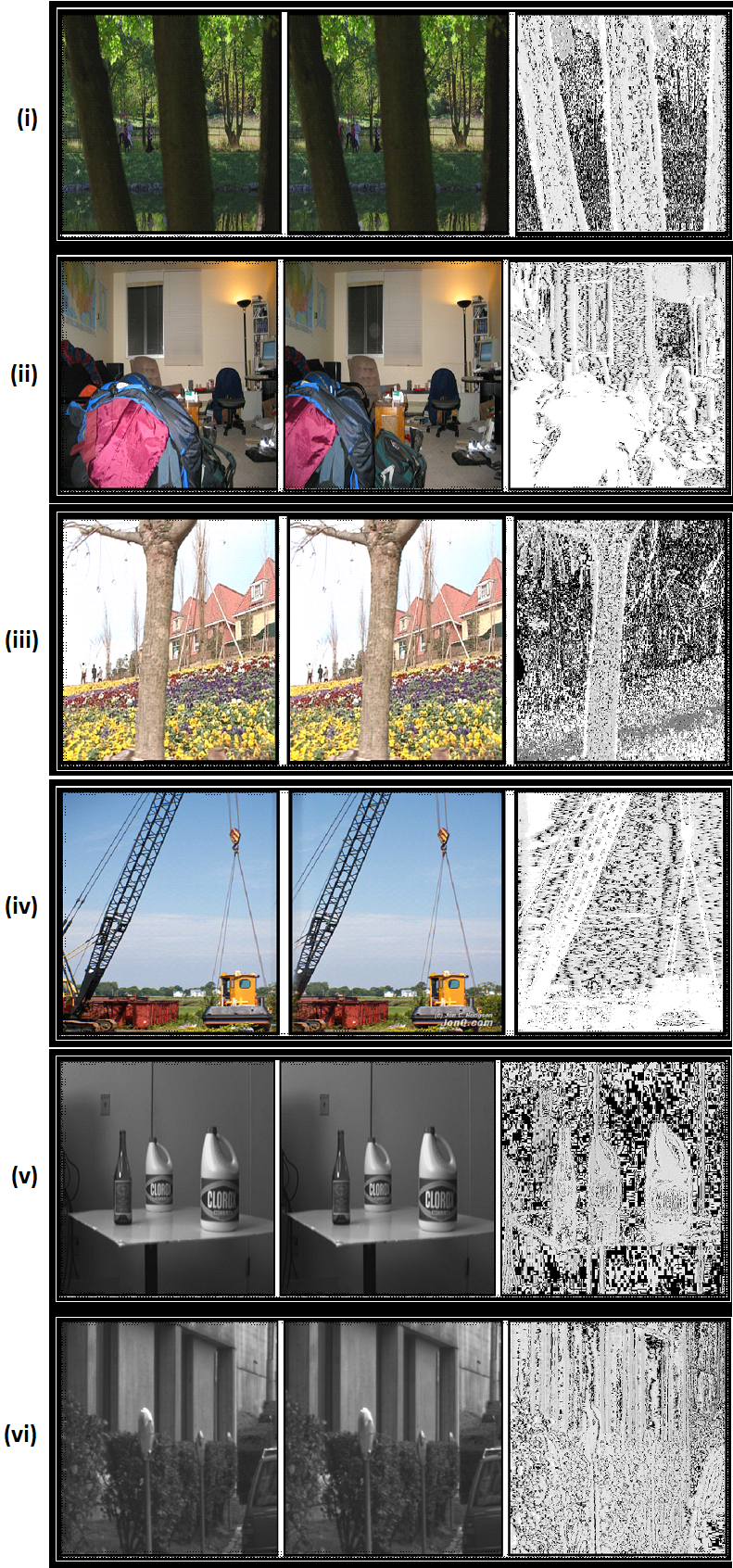}
	\end{center}
	\caption{(a) Left Image; (b) Right Image; (c) Obtained Depth-Map; for different sets of stereoscopic images}
	\label{fig:result}
\end{figure*} 

\begin{figure*}[!htb]
	\begin{center}
		\includegraphics[width=0.98\textwidth]{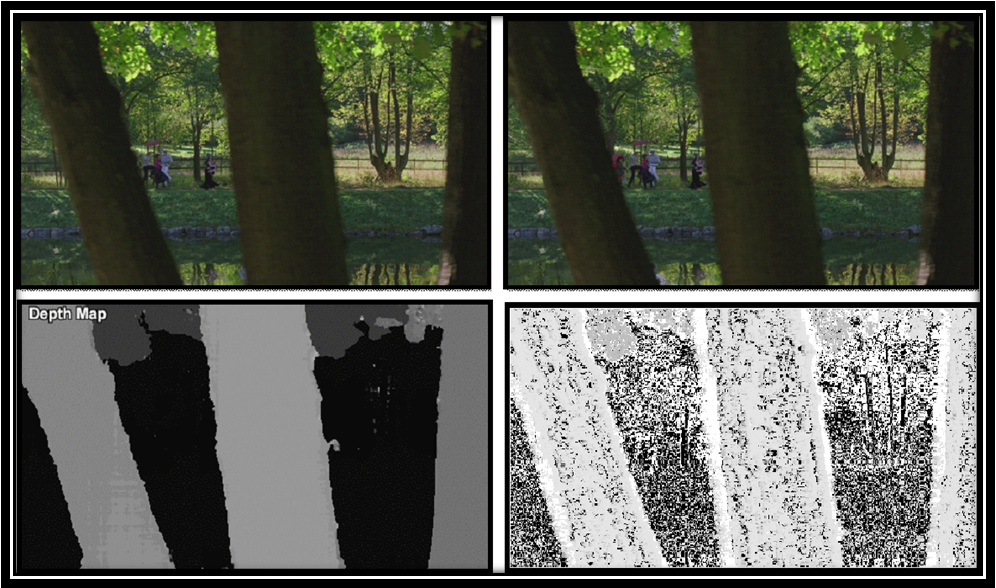}
	\end{center}
	\caption{(a) Left Image; (b) Right Image; (c) Depth-Map by existing algorithms; (d)Depth-Map obtained by proposed algorithm}
	\label{fig:example-depthmap}
\end{figure*} 
% % % % % % % % % % % % % % % % % % % % % %

\subsubsection{Resolving conflicts}
If a situation occurs at which it has been found that some other pixel in right image is similar to that pixel of left image, which is already matched with some other pixel, then the ambiguity is resolved by comparing the Tolerance of previous pixel with that of present pixel. If the tolerance value of present pixel is more then ignore it, otherwise discard the previous pixel and consider present pixel to be the right choice.

\subsubsection{Disparity Calculation}
The method of Binocular disparity is used for creating the depth map. Binocular disparity is the difference between the two images or two eyes, as illustrated in Fig.~\ref{fig:disparity}. Here,

\begin{itemize}
	\item pl : pixel value in the left image
	\item pr: pixel value in the right image corresponding to a similar pixel in left image
	\item f : focal length of the camera
	\item T: difference between the origins of the two cameras
	\item Z: depth value
	\item x$_{l}$: distance of pixel in the left image
	\item x$_{r}$: distance of pixel in the right image
\end{itemize}

The disparity value of a point is often interpreted as the inverse distances to the observed objects. In other words, disparity is inversely proportional to Depth. Therefore, finding the disparity is essential for the construction of the depth map.

\begin{equation} \label{eqn:disparity-1}
Disparity = \frac{fT}{z} = x_{r} - x_{l}
\end{equation}
\begin{equation} \label{eqn:disparity-2}
z = \frac{fT}{\left| x_{l} - x_{r} \right|}
\end{equation}

\subsection{Calculate Pixel value}
For the calculation of color values first calculate the maximum depth from the depth of among all pixels. Give value equal to 0 \textit{i.e.} black color for the pixel of having maximum depth. Then Color value for each pixel is calculated by:

\begin{equation} \label{eqn:color-1}
Color=255-\frac{(depth*255)}{maxdepth}
\end{equation}

Assign value equal to 255 \textit{i.e.} white, to those pixels in left image which do not have any match in right image. By giving these pixel values, a depth-map can be created. In this depth-map darker regions represent that object is far away and lighter regions represent that object is closer to the user. Fig.~\ref{fig:algo-depthmap} depicts the details of the proposed algorithm.

% % % % % % % % % % % % % % % % % % % % % %
\section{Results}
\label{sec:results}

The algorithm has been tested over a large set of images. Few images and their depth maps are shown in Fig.~\ref{fig:result} on the next page. It can be seen that in Fig.~\ref{fig:result}\textit{(i)} trees are nearer to the camera so they are brighter and the ground including background trees are comparatively darker in color. Similarly in the view of a room shown in Fig.~\ref{fig:result}\textit{(ii)} clothes and bag are nearer to the camera so they are white in color and window, rack, chair \& wall have more gray color pixels. In Fig.~\ref{fig:result}\textit{(iii)} darkness increases with the depth \textit{i.e.} from tree to farm house, as the distance from camera increases. The sky in Fig.~\ref{fig:result}\textit{(iv)} is almost black while pulley and other objects are gray and white. The three bottles in Fig.~\ref{fig:result}\textit{(v)} are of different color with respect to their position and the wall appears to be black. The car is the nearest object in Fig.~\ref{fig:result}\textit{(vi)} so it is white as compare to wall of house and small trees,  as they are dark in color.

An example for comparison with an existing algorithm \cite{vatolin} is shown in Fig.~\ref{fig:example-depthmap}. It has two depth-maps: first one is obtained from existing algorithms, and second depth-map is generated by the algorithm proposed in this paper. It can be seen that first depth-map is not much informative, as the places which are far in image appear only as outlines with no clear edges.

\section{Conclusion}
\label{sec:conclusion}
An algorithm was presented to create a depth-map from a pair of stereoscopic (left and right) images.  In a manner different from various known algorithms, the proposed technique matches both images pixel-to-pixel by comparing the RGB components incorporating a tolerance of up to {$\pm$}2.5{\%} in order to handle any kind of dissimilarity which may creep in due to random disturbances, light effects \textit{etc.} in capturing images. The obtained depth-maps were shown to have improvements over the reported ones.

\bibliography{references}
\bibliographystyle{unsrt}

\end{document}